\documentclass[letterpaper, 10 pt, conference]{ieeeconf}  

\IEEEoverridecommandlockouts                              

\usepackage{graphicx}
\usepackage{float}  
\usepackage{color}
\usepackage{amsmath}
\usepackage{booktabs}
\usepackage[colorlinks=true, linkcolor=blue]{hyperref}
\usepackage{cleveref}
\usepackage{multirow}

\title{\LARGE \bf
RALAD: Bridging the Real-to-Sim Domain Gap in Autonomous Driving with Retrieval-Augmented Learning
}

\begin{document}

\author{Jiacheng Zuo$^{*,1}$,  Haibo Hu$^{*,2}$, Zikang Zhou$^{2}$, Yufei Cui$^{3}$, Ziquan Liu$^{4}$,\\
Jianping Wang$^{2}$, Nan Guan$^{2}$, Jin Wang$^{\dag,1}$, Chun Jason Xue$^{5}$ 
\thanks{*Equal contribution \qquad \dag Corresponding author}
\thanks{$^{1}$Department of Future Science and Engineering, Soochow University.
        {\tt\small jczuo102137@stu.suda.edu.cn, wjin1985@suda.edu.cn}}%
\thanks{$^{2}$Department of Computer Science, City University of Hong Kong.
        {\tt\small \{haibohu2-c, zikanzhou2-c\}@my.cityu.edu.hk,}
        {\tt\small \{nanguan, jianwang\}@cityu.edu.hk}}%
\thanks{$^{3}$Department of Computer Science,
        McGill University.
        {\tt\small yufeicui92@gmail.com}}%
\thanks{$^{4}$School of Electronic Engineering and Computer Science,
        Queen Mary University of London.
        {\tt\small ziquan.liu@qmul.ac.uk}}%
\thanks{$^{5}$Department of Computer Science,
        Mohamed bin Zayed University of Artificial Intelligence.
        {\tt\small jason.xue@mbzuai.ac.ae}}%
}

\maketitle
\thispagestyle{empty}
\pagestyle{empty}

\begin{abstract}
As end-to-end autonomous driving advances toward real-world deployment, ensuring the safety of autonomous vehicles (AVs) has become a critical requirement for their commercial viability. While rule-based AVs have traditionally undergone rigorous testing in both real-world and simulated environments before deployment, data-driven autonomous models are typically trained on real-world datasets, limiting their generalization to simulation environments. This poses a significant challenge for the development and testing of end-to-end autonomous driving. To address this issue, we propose Retrieval-Augmented Learning for Autonomous Driving (RALAD), a novel framework designed to bridge the real-to-sim gap in a cost-effective manner. RALAD consists of three key components: (1) domain adaptation via an enhanced Optimal Transport (OT) method, which retrieves the most similar scenarios between real and simulated environments; (2) feature fusion across similar scenarios, enabling the construction of a feature mapping between real-world and simulated domains; and (3) feature extraction freezing with fine-tuning on the fused features, allowing the model to learn simulation-specific characteristics through feature mapping. We evaluate RALAD on three monocular 3D object detection models, and the results demonstrate that our approach significantly improves model accuracy in simulation. Additionally, we use real autonomous vehicle for testing in real-world scenarios, and have established simulated scenes similar to reality for further testing, which illustrate the effectiveness of our method. 
\end{abstract}

\section{INTRODUCTION}
As the application of machine learning in autonomous driving continues to gain unstoppable momentum~\cite{c1}, a vast array of models has emerged for solving various autonomous driving tasks~\cite{c2,c3,c4}, including image segmentation~\cite{c5,c6,c7}, object detection~\cite{c8,c9}, and motion planning~\cite{c10}. These models are usually trained and tested using real-world datasets such as KITTI~\cite{c11}, Waymo~\cite{c12}, and nuScenes~\cite{c13}, which cover common driving scenarios.
\begin{figure}[thpb]
  \centering
  \includegraphics[width=0.48\textwidth]{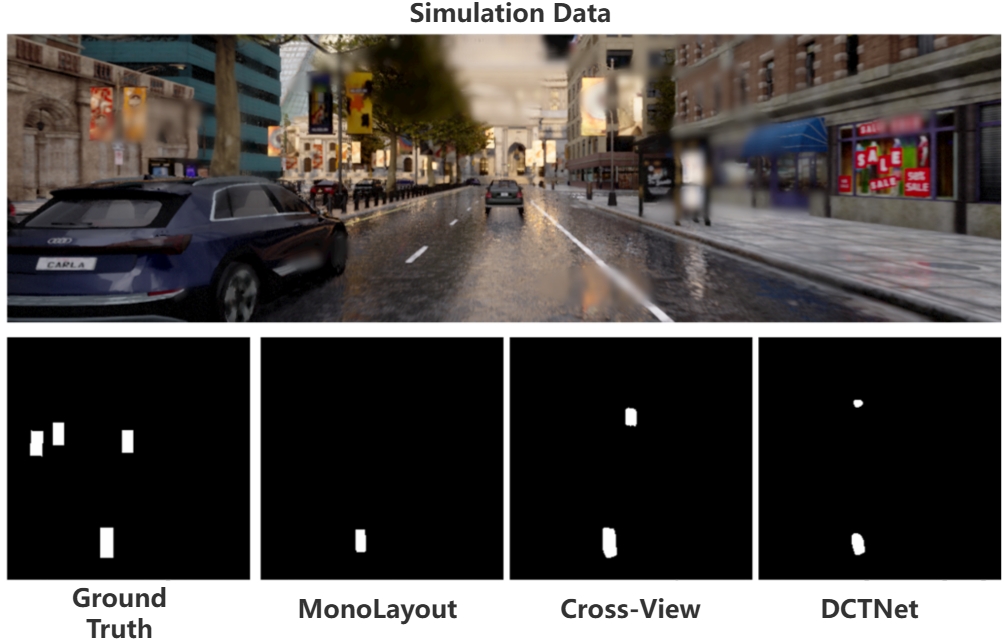}
  \vspace{-20pt}
  \caption{\textbf{Performance of baseline model in simulated rainy weather:} All models showed varying degrees of performance degradation under simulated rainy weather conditions.}
  \vspace{-20pt} 
  \label{fig:bad show}
\end{figure}
However, autonomous driving systems inevitably encounter corner cases, such as extreme weather conditions, unexpected pedestrian behavior, and rare road scenarios, which challenge their perception and decision-making capabilities~\cite{c14,c15}. Given the high safety standards required in vehicle operation to prevent life-threatening accidents, it is imperative that autonomous systems rigorously address and test these corner cases to ensure robust performance~\cite{c14, c15}. 

Models trained on real-world datasets predominantly encounter regular driving scenarios, making it challenging to cover and address the rare and complex corner cases~\cite{c21,c16,c19}. To replicate and test corner cases, simulators such as CARLA and AirSim are commonly used to replicate and test specific scenarios. This approach has proven highly effective for rule-based autonomous driving in the past.~\cite{c21}. However, with the rise of end-to-end autonomous driving, vision-based models impose significantly different granularity requirements on simulated and real-world scenes, resulting in a substantial domain gap. As shown in Figure \ref{fig:bad show}, we collected a simulated dataset from CARLA and evaluated three models. All three models exhibited a performance decrease when tested on simulation, indicating that although these models perform well in the real-world, their performance tends to degrade in the simulated environment.

Current approaches to addressing this issue typically involve training the model directly on a mixture of real and simulated datasets~\cite{modelbysim} to mitigate the domain gap. However, these methods require full model retraining, incurring computational costs while potentially compromising the model's original performance. Therefore, developing an efficient and minimally intrusive solution remains a significant challenge. While there are numerous domain generalization methods, many still require significant adjustments to the model. In contrast, retrieval-augmented learning (RAL) offers a more lightweight approach by enhancing the model's ability to generalize across domains without the need for extensive retraining. In particular, methods using nearest neighbor search and feature fusion for domain adaptation~\cite{c16,c18} inspire our approach to leveraging RAL to bridge the cross-domain gap between real and simulated datasets in autonomous driving.

While RAL has shown promise in domains like computer vision, applying this technique to autonomous driving introduces several challenges. Firstly, RAL was originally developed for classification tasks, whereas autonomous driving tasks often involve more complex, high-dimensional tasks like 3D detection, trajectory prediction, and decision-making, this creates a challenge in adapting RAL to these tasks. Furthermore, previous applications of RAL typically involved datasets from the same domain but different fields, such as Camelyon16 and Camelyon17. In contrast, real and simulated data in autonomous driving are fundamentally different in dimensions, which presents an additional challenge. Lastly, due to the sheer volume of autonomous driving data, efficiently adjust the model without affecting its original feature representations is a significant challenge.

We propose RALAD to address the gap between real and simulated images in autonomous driving. We introduce an enhanced Optimal Transport (OT) method for domain adaptation, which retrieves the most similar scenarios between real and simulated environments, and then through feature fusion, we construct a feature mapping between real and simulated feature maps. Subsequently, we utilize this feature mapping to allow the model to infer real-world features from simulated features. This approach eliminates the need for the model to learn features from simulated data, instead focusing on establishing the mapping between simulated and real features. Additionally, we freeze the feature extraction module, preserving the model’s original performance while significantly reducing the computational cost of training.

Our contributions are summarized as follows:
\begin{itemize}
\item We introduce RALAD, a framework utilizing OT method to establish a mapping between simulation and reality, in order to address the gap between reality and simulation in autonomous driving. 
\item We apply RALAD to three models, achieving an improvement of up to 11.02\% in accuracy on simulation while maintaining the accuracy on KITTI. 
\item We use real autonomous vehicles and CARLA to create real and simulated environments, and conduct extensive experiments to validate the RALAD. 
\end{itemize}

    \begin{figure*}[!t]
    \centering
    \includegraphics[width=0.9\textwidth, height=0.35\textheight]{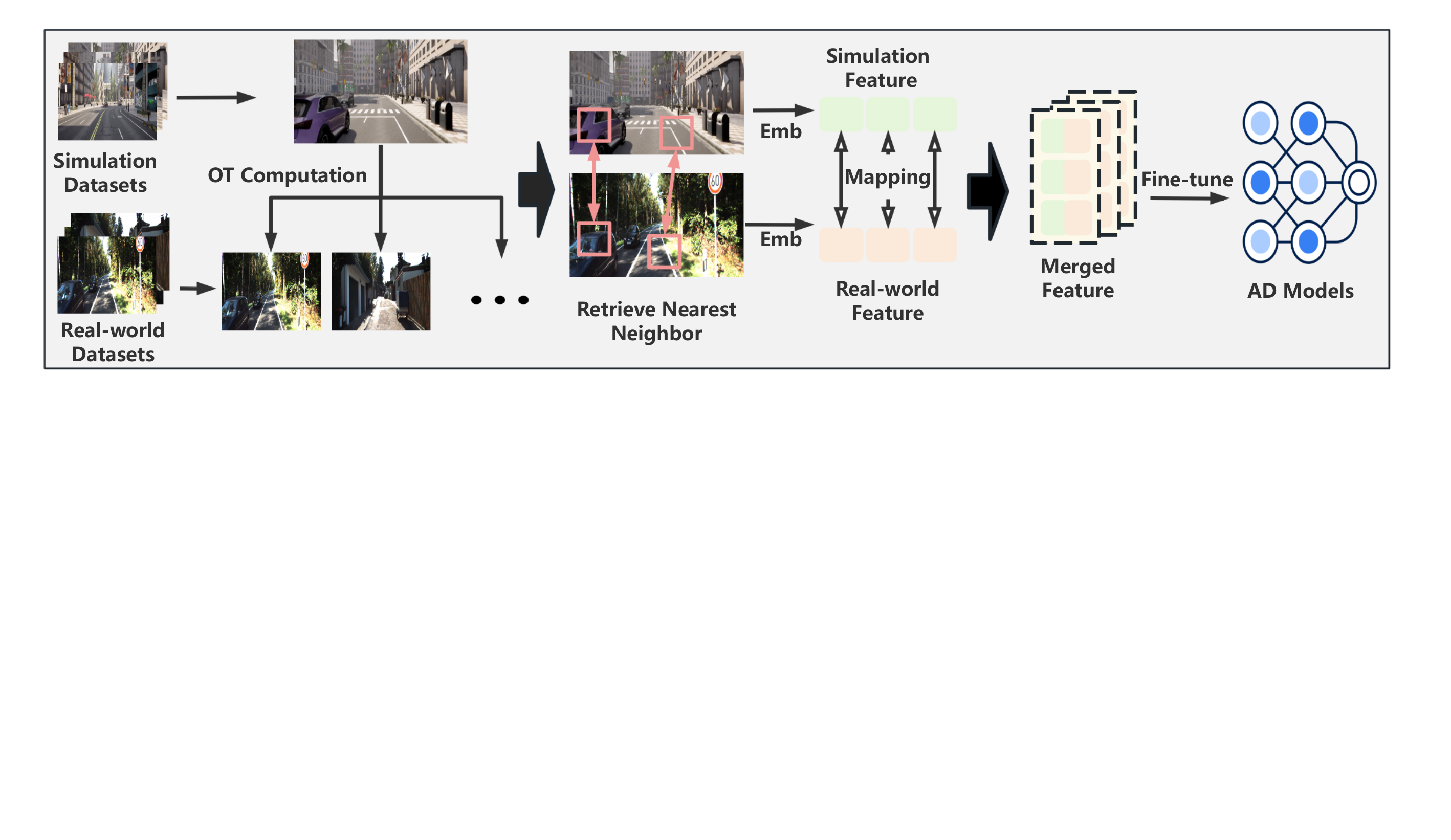}
    \vspace{-140pt}
    \caption{Overview of the RALAD framework for autonomous driving. The process begins with simulation and real-world datasets, which undergo Optimal Transport (OT) computation to retrieve the real feature map with the smallest transfer distance for the simulated feature map. The features from both domains are then mapped and merged, followed by fine-tuning on the combined features to enhance the model's performance on autonomous driving (AD) tasks.}
    \vspace{-10pt}
    \label{fig:cross-view+ram}
    \end{figure*}
    
In the following, we discuss the related work in Section \ref{sec:related_work} and detail the RALAD framework in Section \ref{sec:method}, with Section \ref{sec:experimental} presenting the experimental results and Section \ref{sec:conclusions} concluding the paper.

\section{RELATED WORK}
\label{sec:related_work}
    \subsection{Retrieval-Augmented Learning} RAL is an approach that enhances learning models, it integrates retrieval mechanism to leverage existing data representations, thereby improving performance and efficiency. For example, Yottixel~\cite{c22} employs a mix of supervised and unsupervised methods, including segmentation, clustering, and deep networks, to analyze image patches and employ distance metrics for retrieval. SISH~\cite{c23} utilizes a tree structure for rapid WSI search and an uncertainty-based ranking for retrieval, reducing storage and labeling by building on preprocessed mosaics. HHOT~\cite{c24} introduces optimal transport (OT) as a metric for comparing whole slide images (WSIs) or across WSI datasets, theoretically underpinning the application of OT for steering the retrieval and assembly of datasets. In RAM-MIL~\cite{c18}, the attention weight serves as a measure of probability density, signifying the "mass" being transferred. By quantifying this, the method computes the conversion cost across various data domains. It then employs this distribution for nearest-neighbor retrieval, seamlessly integrating features from distinct domains to address out-of-domain challenges.
    
    \subsection{Gap Between Real And Sim}
    In the field of autonomous driving, there exists a gap between simulation and reality~\cite{c25}, which is caused by discrepancies in lighting, textures, vehicle dynamics, and agent behaviors. To address this, researchers have developed two primary approaches: knowledge transfer learning and digital twins (DTs)~\cite{c17,c28,c30}. In knowledge transfer learning, the reality gap problem is compounded by uneven environmental sampling and complex physical parameters~\cite{c26}. To overcome this, researchers have developed strategies such as curriculum learning, meta-learning, and domain randomization~\cite{c27}. Domain randomization, in particular, helps align simulation parameters with real-world variability, facilitating the transfer of learned strategies to real-world applications. Conversely, digital twin technology creates virtual models of real-world entities or systems. A case in point is the SynFog dataset~\cite{c28}, which uses an end-to-end process to produce photo-realistic synthetic data. Nonetheless, despite their promise, these methods confront the issue of high computational expenses, particularly in complex, dynamic real-world settings.

     \subsection{3D Object Detection in Autonomous Driving}
     3D object detection is crucial for autonomous driving as it enables vehicles to accurately perceive their surroundings, which is essential for safe navigation and decision-making. Conventionally, this has been achieved with the help of LiDAR sensors, which, although precise, are prohibitively expensive and computationally demanding. To address these limitations, the field has seen a significant advancement with the application of deep learning that leverage cameras for 3D detection. Specifically, the development of bird’s-eye view (BEV) representations from monocular images has emerged as a promising and more cost-effective alternative. Techniques such as MonoLayout~\cite{c5} and CrossView~\cite{c6} demonstrate the potential of using these BEV to perform 3D object detection. CrossView, in particular, has introduced a cross-view transformation module and a context-aware discriminator to enhance results, achieving cutting-edge performance in vehicle occupancy estimation. Nonetheless, issues such as class imbalance and low computational efficiency remain. The Dual-Cycled Cross-View Transformer network (DctNet)~\cite{c7} has been proposed to tackle these challenges by integrating focal loss and optimizing multi-class learning, setting new standards of performance in the field of 3D object detection.

\section{METHODOLOGY}
 \label{sec:method}   
    This section provides a detailed explanation of the RALAD framework, as shown in Figure~\ref{fig:cross-view+ram}. We introduce Retrieval-Augmented Learning based on Optimal Transport into autonomous driving.
    
    \subsection{Problem Formulation}
    The gap between real-world and simulation (real2sim) primarily arises from the cross-domain challenges between real and simulated data in autonomous driving. To address this, we consider two datasets: the Real dataset $D_r$ and the Sim dataset $D_s$. The Real dataset is defined as $D_r=\left\{X_n, Y_n\right\}_{n=1}^{N_r}$, where $N_r$ is the number of images, and the Sim dataset is $D_s=\left\{\tilde{X_m},\tilde{Y_m}\right\}_{m=1}^{{N_s}}$, with ${N_s}$ being the number of simulated images, X and Y represent the image and ground truth, respectively. To bridge the gap between these two datasets, the first step is to establish a mapping relationship between them. We extract features from real and simulated images using an encoder function g(·), resulting in $H_k$ = g($X_k$) for real images and $\tilde{H}_k$ = g($\tilde{X}_k$) for simulated images. 
    Our ultimate goal is to create mapping relationship between features from different domain, thereby find the most similar real feature map from the real dataset for each simulated feature map. The formula is as follows:
    \begin{equation}
        \begin{aligned}
        \mathcal{L}(\tilde{H}_i, D_r) = \min_{j=1}^{N_r} \operatorname{d_{OT}}\left(\tilde{H}_i,H_j\right)
        \end{aligned}
    \end{equation}
    \subsection{Retrieval-Augmented Learning based on Optimal Transport in Real2Sim}
\begin{figure}[thpb]
      \centering
      \includegraphics[width=0.48\textwidth]{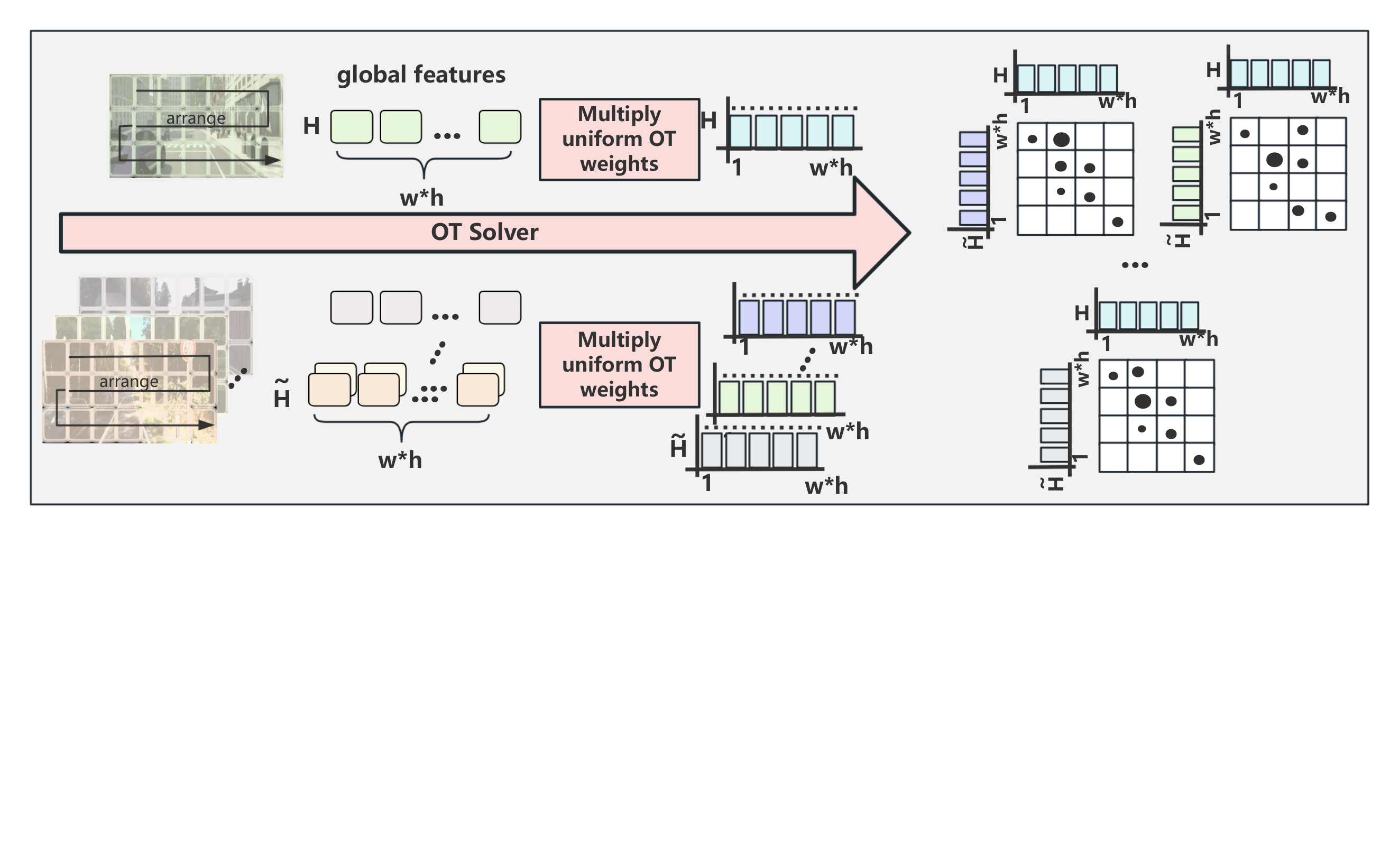}
      \vspace{-70pt}
      \caption{The one round calculation process of Optimal transport: the simulated image features of the single frame above and all the real image features below are separated, then each feature is multiplied by the same OT weights, and finally the simulated features of the single frame and all the real features are calculated for Optimal transport one by one. }
      \vspace{-5pt} 
      \label{OTfigure}
    \end{figure}

\begin{figure*}[thpb]
        \centering
        \includegraphics[height=0.15\textheight, width=0.9\textwidth]{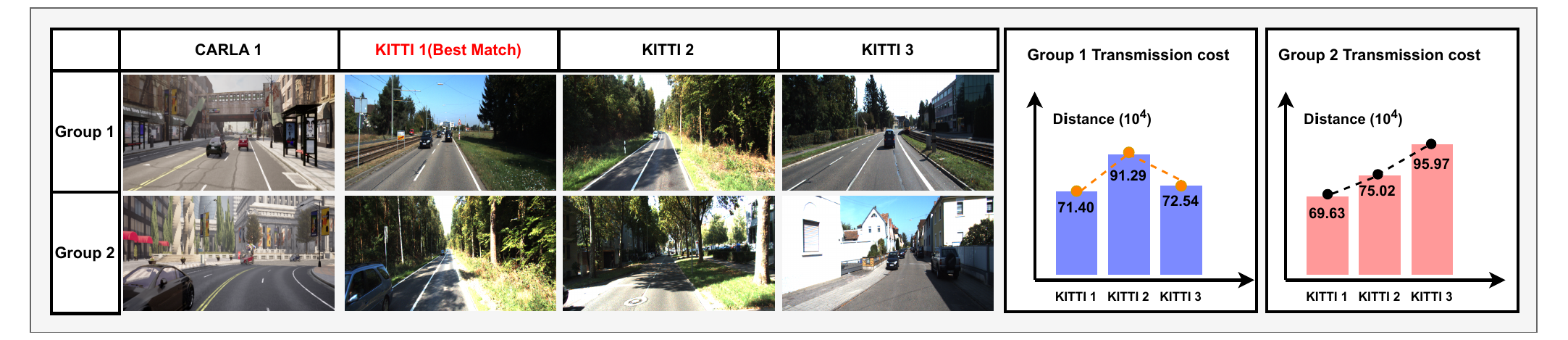}
        \vspace{0pt}
        \caption{The table displays two sets of nearest neighbor results for KITTI 1, with these results drawn from the CARLA dataset. CARLA 1 stands out as the optimal match for KITTI 1 among the retrieved neighbors. And the table, a pair of bar graphs illustrate the transmission cost comparison between KITTI 1 and the CARLA features, a lower transmission cost indicates a higher degree of similarity between the images.}
        \vspace{-10pt}
        \label{fig: OT_distance}
    \end{figure*}

    To address the mapping relationship between features from different domains, we introduce global feature level Optimal Transport, because the previous RAL method was mainly local and focused on the differences between individual instances in high-definition medical images, the instance images were relatively clear and did not have much interference.  However, in autonomous driving, the environment has a significant impact on vehicle recognition, such as in rainy or snowy weather. So we proposed a global feature level Optimal Transport to globally quantify the differences between reality and simulation. As shown in Figure \ref{OTfigure}, we use the encoder of the pre-trained model to extract features from a single simulated image and all real images, and then flatten the feature map to generate row vectors. Then the simulated vectors are subjected to OT calculations with all real vectors. This ensures that a distribution correlation can be established between simulated and real features, enabling the discovery of more optimal real features, as shown in Figure \ref{fig: OT_distance}, where vehicles and lanes almost correspond. This mapping process is repeated on the simulated dataset until the traversal of the simulated dataset is completed.
    
    
    During the OT calculation process, each image is treated as a probability distribution. We assign uniform weights to all instances and leave the non-uniform OT calculation to future exploration. The purpose of the OT algorithm is to calculate the distance between two features $H_n$ (real-world) and $\tilde{H_m}$ (simulation) for retrieval of the nearest feature. The formula is as follows:
    \begin{equation}
        \label{OT}
        \begin{aligned}
        \operatorname{d_{OT}}\left(H_n, \tilde{H_m}\right)=\sum_{i=1}^{w \cdot h} c\left(h_{ni}, \tilde{h_{mi}}\right) T_{i i}+\beta \cdot \sum_i T_{i i} \log T_{i i} \\
        \text { s.t. } T^T 1_{w \cdot h}=H_n, T1_{w \cdot h}=\tilde{H}_n, T \geq 0\,.
        \end{aligned}
    \end{equation}
    
    In this equation, T denotes the transport plan matrix where each element $T_{ii}$ specifies the amount of "mass" to be transported from $h_{ni}$ to $\tilde{h_{mi}}$, $h_{ni}$  and $\tilde{h_{mi}}$ are a feature value located at the real and simulated feature maps, respectively, w and h are the width and height of the feature map. The function c($h_{ni}$, $\tilde{h_{mi}}$) is a cost function that quantifies the cost of transporting a unit of mass from $h_{ni}$ to $\tilde{h_{mi}}$. A common choice of c($h_{ni}$, $\tilde{h_{mj}}$) is the squared $l_2$ distance between the features, i.e., c($h_{ni}$, $\tilde{h_{mi}}$) = $||h_{ni} - \tilde{h_{mi}}||_2^2$, here, $1_{w \cdot h}$ is vector of
    ones. We also introduced entropy regularization~\cite{c29} to reduce sensitivity to outlier instances.

    \subsection{Convex Merge and Fine-Tune}
    In the feature retrieval phase of autonomous driving, we have already obtained it through OT computed the optimal matching simulated feature $\tilde{H}^*$. Upon finding this match, we employ a convex merge operation to combine the real feature $H$ with the simulated feature $\tilde{H}^*$, resulting in a new composite feature $\hat{H}$, calculated as $\hat{H}=\pi(H, \tilde{H}^*)$, as shown in Figure \ref{fig:cross-view+ram}. The merging function $\pi$(·) is typically a convex combination method, convex fusion of features H and ${H}^*$ is achieved through complementary weight coefficients, where the coefficients ensure that the fusion result is located on the line segment formed by the two features, achieving linear interpolation and balance in the feature space, this fusion also allows us to establish mapping relationships on features.
    The set of merged features denoted as $D_m=\{\hat{H_i}\}^{N_m}_{i=1}$, with $N_m$ being the number of features, is then utilized for the fine-tuning phase. 

\begin{figure}[thpb]
      \centering
      \includegraphics[width=0.48\textwidth]{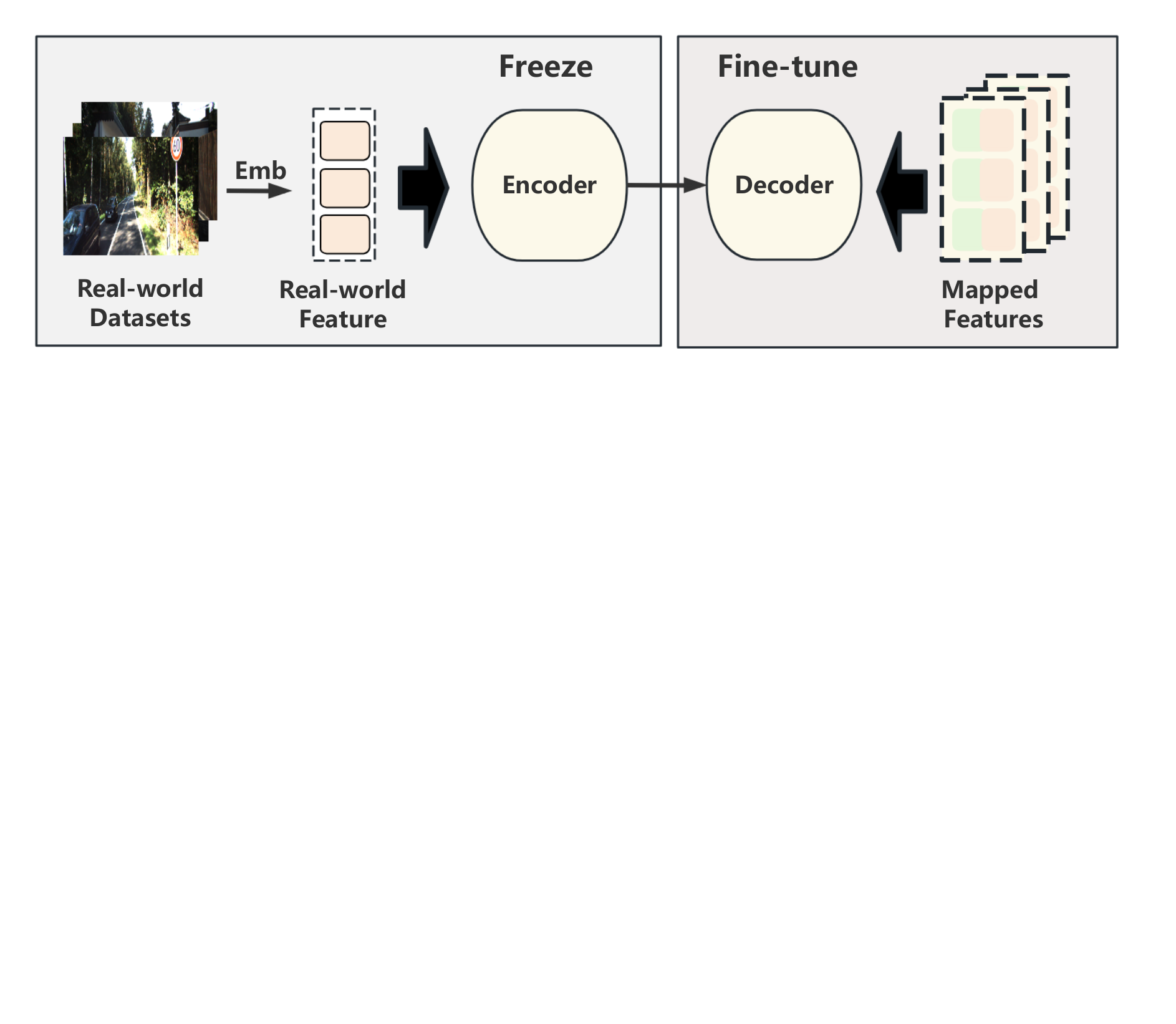}
      \vspace{-150pt}
      \caption{The process of fine-tune: the model is frozen except for the decoder, and merged features are as inputs for the decoder to learn the mapping relationship between simulation and reality.}
      \vspace{-10pt} 
      \label{FineTunefigure}
    \end{figure}

    During fine-tuning, we maintain the integrity of the pre-trained model by freezing all layers except for the decoder, because we previously used an encoder to extract features, then combined OT and Convex Merge to extract the mapping between real and simulated features, in order to maintain the ability to extract the mapping between real and simulated features, we froze the encoder layer of the model. Then, through this extracted mapping relationship, we only need to train the decoder layer of the model, so that the model can learn to establish a mapping relationship between reality and simulation, as shown in Figure~\ref{FineTunefigure}.
    
    \section{EXPERIMENT}
    \label{sec:experimental}
    To evaluate the effectiveness of RALAD , we conduct experiments on three models trained on the KITTI dataset, a widely recognized benchmark in autonomous driving. We first present the metrics and dataset details, followed by experimental results showcasing the improvements achieved by RALAD, then we analyze the training efficiency of RALAD and provide a qualitative visualization on one of the models. Finally, we conduct a series of ablation studies to further validate our approach.
    \subsection{Dataset and Metrics}
     The workstation used for this task was equipped with a single NVIDIA RTX A4000 GPU card. All the input images are normalized to $1024 \times 1024$ and the output size is $256 \times256$. The network parameters are randomly initialized and we adopt the Adam optimizer~\cite{c29} and use a mini-batch size of 6. The initial learning rate is set to $1 \times 10^{-4}$, and it is decayed by 0.1 after 25 epochs.

    The KITTI dataset comprises 7481 monocular images from vehicle front cameras, split into 3712 training and 3769 validation images based on \cite{c20}, with ground truth derived from~\cite{c5}. For the CARLA dataset, collected via the CARLA 0.9.15 software with robust annotation features, we gathered data from maps like Town01, Town02, and Town07, comprising 500 training, 473 validation, and 227 test images, totaling 1200. Utilizing the RALAD, we extracted 4066 features from the KITTI and CARLA training sets and fine-tuned the models on KITTI, the features were allocated to training (2536) and validation (1530) sets. We then assessed the fine-tuned models on both datasets to showcase its real-world and simulated performance, using Mean Intersection over Union (mIOU) and Mean Precision (mAP) as metrics.

    \subsection{Experimental Results}  
    
    Our experimental metrics are presented in Table \ref{table:ablation_study_combined_single_column}, demonstrating the effectiveness of RALAD across all three models. RALAD not only preserves the models’ accuracy on real-world data but also significantly improves their performance on simulated data, achieving recognition capabilities comparable to or even exceeding those on real data. Furthermore, we observe that RALAD provides greater performance gains for newer models, highlighting its effectiveness in enhancing state-of-the-art architectures.

    \textbf{MonoLayout:} RALAD shows minimal improvement on the KITTI dataset (mIOU from 30.18\% to 30.26\%, with a slight mAP decrease from 45.91\% to 44.98\%). However, on the CARLA dataset, it achieves significant improvements (mIOU from 25.26\% to 34.13\%, mAP from 48.93\% to 56.41\%). This suggests that RALAD helps more in simulation environments, particularly for this model.

    \textbf{CrossView:} the gains on KITTI are modest (mIOU from 38.85\% to 39.21\%, mAP from 56.64\% to 56.49\%), while on CARLA, the improvements are much more pronounced (mIOU from 30.55\% to 40.82\%, mAP from 53.25\% to 65.54\%).

    \textbf{DctNet:} the improvements on KITTI are slight (mIOU from 39.44\% to 39.07\%, mAP from 58.89\% to 57.64\%), but the model benefits greatly from RALAD on CARLA (mIOU from 31.09\% to 42.11\%, mAP from 54.72\% to 67.24\%). These results suggest that RALAD has a more substantial impact on the simulation dataset, improving both detection accuracy and precision.
    \begin{table}[!tb]
    \centering
    \caption{Model Performance on KITTI and CARLA}
    \label{table:ablation_study_combined_single_column}
    \renewcommand{\arraystretch}{1.1} 
    \setlength{\tabcolsep}{0.5pt} 
    \begin{tabular}{|c|c|c|c|c|}
        \hline
        \multirow{2}{*}{Methods} & \multicolumn{2}{c|}{KITTI} & \multicolumn{2}{c|}{CARLA} \\
        \cline{2-5}
        & mIOU(\%) & mAP(\%) & mIOU(\%) & mAP(\%) \\
        \hline
        MonoLayout~\cite{c5} & 30.18 & 45.91 & 25.26 & 48.93 \\
        MonoLayout+RALAD (ours) & 30.26 & 44.98 & \textbf{34.13}$\uparrow$ & \textbf{56.41}$\uparrow$ \\
        \hline
        Cross\_view + $l_2$ & 37.29 & 54.29 & 34.28 & 53.27 \\
        Cross\_view~\cite{c6} & 38.85 & 56.64 & 30.55 & 53.25 \\
        Cross\_view+RALAD (ours) & 39.21 & 56.49 & \textbf{40.82}$\uparrow$ & \textbf{65.54}$\uparrow$ \\
        \hline
        DctNet + $l_2$ & 39.05 & 60.69 & 37.89 & 61.15 \\
        DctNet~\cite{c7} & 39.44 & 58.89 & 31.09 & 54.72 \\
        DctNet+RALAD (ours) & 39.07 & 57.64 & \textbf{42.11}$\uparrow$ & \textbf{67.24}$\uparrow$ \\
        \hline
    \end{tabular}
    \vspace{-2mm}
    \end{table}
    
    \begin{table}[!tb]
    \centering
    \caption{Training Cost}
    \label{table:ablation_study_time}
    \renewcommand{\arraystretch}{1.1} 
    \setlength{\tabcolsep}{9pt} 
    \begin{tabular}{|c|c|c|}
        \hline
        Methods & KITTI & CARLA \\
        \hline
        MonoLayout~\cite{c5} & 34.84 s/epoch & 33.97 s/epoch \\
        MonoLayout+RALAD & 3.81 s/epoch & 4.07 s/epoch \\
        \hline
        Cross\_view~\cite{c6} & 41.54 s/epoch & 41.33 s/epoch \\
        Cross\_view+RALAD & 5.27 s/epoch & 5.29 s/epoch \\
        \hline
        DctNet~\cite{c7} & 68.07 s/epoch & 67.97 s/epoch \\
        DctNet+RALAD & 20.23 s/epoch & 21.08 s/epoch \\
        \hline
    \end{tabular}
    \end{table}  
    \subsection{Training Overhead} Our RALAD adopts the fine-tuning approach, which can reduce the time cost of re-training existing models, as shown in Table \ref{table:ablation_study_time}. To compare the performance of RALAD fine-tuning with standard model re-training, we conducted experiments using 1200 features in a consistent environment. For CrossView, the original model requires 41.54 seconds per epoch on KITTI and 41.33 seconds on CARLA in re-train, while RALAD fine-tuning only needs 5.27 seconds on KITTI and 5.29 seconds on CARLA. In comparison, MonoLayout and DctNet also show significant reductions in training time with RALAD, from 34.84 to 3.81 seconds on KITTI and from 68.07 to 20.23 seconds, respectively. These results highlight RALAD's efficiency, especially for large-scale adaptation.

    \subsection{Visualizations}
    \label{sec:visual}
    \textbf{Visualization of Retrival Results:}
     As shown in Figure \ref{fig: OT_distance}, we visualized the nearest sets of feature maps retrieved by RALAD, to verify that our method can retrieve highly approximate simulated and real-world feature maps. It can be seen that the method successfully retrieves the most approximate feature map from each dataset. Specifically, a lower OT distance indicates a higher degree of similarity between two feature maps, such as those from KITTI1 and CARLA1. Both groups demonstrate high consistency in key features (e.g., vehicle positions and lane conditions) and have the minimum transmission costs of 62.27$ \times 10^4$ and 70.21$ \times 10^4$, respectively. These results highlight the capability of RALAD to accurately compute the similarity between feature maps in practical applications, enabling effective matching of real and virtual data.
    
    \begin{figure}[thpb]
      \centering
      \includegraphics[width=0.4\textwidth, height=0.18\textheight]{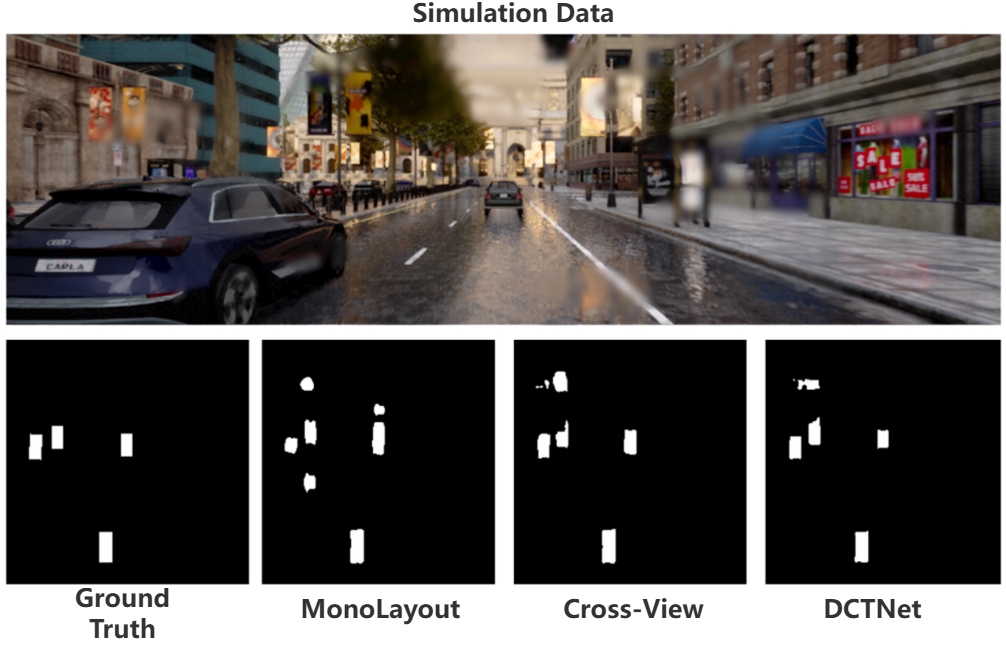}
      \vspace{-10pt}
      \caption{\textbf {The performance of models with RALAD in simulated rainy weather:} All models show varying degrees of performance improvement under simulated rainy weather conditions.}
      \vspace{-10pt} 
      \label{figurelable2}
    \end{figure}
        \begin{figure}[thpb]
        \centering
        \includegraphics[width=0.4\textwidth, height=0.18\textheight]{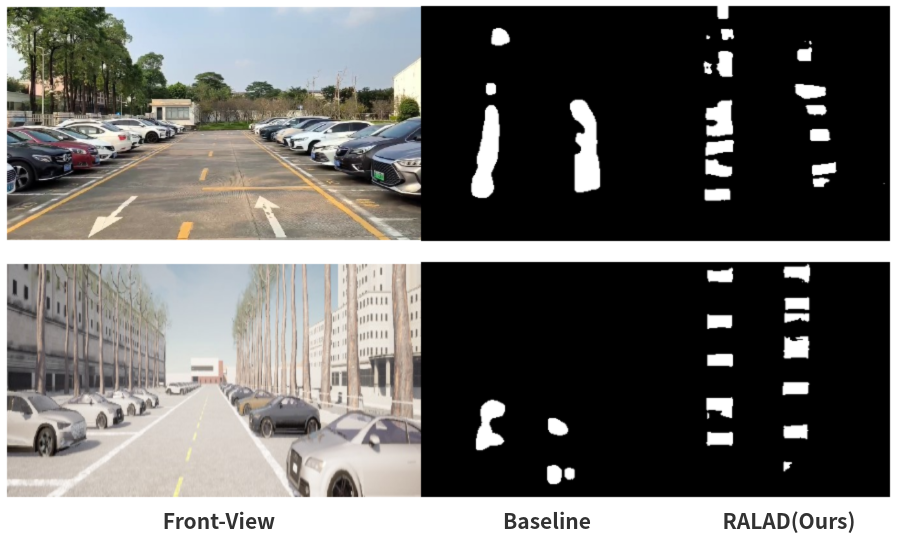}
        \vspace{-10pt}
        \caption{\textbf{Evaluation in the real world:} \textbf{Center}: the compared baseline (Cross\_view). \textbf{Right}: the baseline enhanced by RALAD (RALAD\_Cross\_view).}
        \vspace{-10pt}
        \label{fig: real-world}
    \end{figure}

     \textbf{Simulation and Real Autonomous Vehicle Experiments:} To further demonstrate the improvements of RALAD on CARLA dataset, we re-tested the scene from Figure \ref{fig:bad show}, as shown in Figure \ref{figurelable2}, there is a significant improvement in detecting vehicles in simulation, especially with DctNet, which successfully identifies vehicles entirely. In addition, we collected parking lot data for testing using an autonomous vehicle equipped with the open-source Autoware.Universe autonomous driving system. Considering the unique characteristics of the real and simulated environments, we reconstructed a scene in CARLA that closely resembles the real world to minimize external factors, thereby demonstrating the improvements provided by RALAD compared to the original model. As shown in Figure \ref{fig: real-world}, the models with RALAD provide more accurate detection in both real and simulated environments compared to the original performances.
\begin{table}[!tb]
    \centering
    \caption{Convex Combination of Ratio Setting}
    \label{table:convex_combination_ratios}
    \renewcommand{\arraystretch}{1.1} 
    \setlength{\tabcolsep}{9pt} 
    \begin{tabular}{|c|c|c|}
        \hline
        \multirow{2}{*}{KITTI:CARLA} & \multicolumn{2}{c|}{Metrics (mIOU/mAP)} \\
        \cline{2-3}
        & KITTI & CARLA \\
        \hline
        0.7:0.3 & 34.70 / 54.48 & 22.06 / 47.62 \\
        0.6:0.4 & 32.10 / 54.02 & 35.12 / 63.33 \\
        0.5:0.5 & 30.21 / 52.83 & 35.77 / 60.74 \\
        0.4:0.6 & 29.68 / 49.33 & 36.83 / 61.39 \\
        \hline
    \end{tabular}
    \end{table}
    \subsection{Ablation Experiment} 
    \textbf{OT Ablation:} In this section, we conducted an ablation study on the OT algorithm by separately preserving the $l_2$ distance for experiments. The results are shown in Table \ref{table:ablation_study_combined_single_column}. It can be seen that using the $l_2$ distance to establish a mapping relationship does not show a significant difference compared to the baseline on KITTI, and also demonstrates a improvement on CARLA. However, when compared to using RALAD, the gap is extremely noticeable. The CrossView using $l_2$, as opposed to the OT, has a slight decrease on KITTI (mIOU from 39.21\% to 37.29\%, mAP from 56.49\% to 54.29\%), and a significant decrease on CARLA (mIOU from 40.82\% to 34.28\%, mAP from 65.54\% to 53.27\%).
    And DctNet using $l_2$ also shows a significant decrease on the CARLA dataset (mIOU from 42.11\% to 37.89\%, mAP from 67.24\% to 61.15\%), as opposed to the OT. This experiments clearly indicates that the OT algorithm is more accurate than the $l_2$ algorithm in establishing a mapping relationship between real and simulated environments, enabling the model to better bridge the gap between reality and simulation.

    \textbf{Convex Combination Ratio:}
  In the feature merging stage of RALAD, different fusion ratios can produce varied results for the model. To explore this, we conducted comparison experiments with four different ratio settings, validating them on the KITTI and CARLA datasets, using a total of 1800 features. As shown in Table \ref{table:convex_combination_ratios}, KITTI: CARLA represent the weight coefficients for real and simulated feature maps, respectively, during the merge, the ratio of the convex merge is based on the ~\cite{c18}, and we have also conducted additional validation experiments, which demonstrate that the 0.6:0.4 combination achieved a balanced performance, slightly decreasing on KITTI, but showing a substantial improvement on CARLA, making it an optimal trade-off between real and simulated data. In contrast, the 0.7:0.3 combination performed better on KITTI , but much worse on CARLA. As the CARLA ratio increased, performance on CARLA continued to improve, reaching the highest metrics with the 0.4:0.6 combination, but at the cost of reduced performance on KITTI, so we recommend and adopt this ratio for feature fusion to obtain better results.
\subsection{Experiments Conclusion}
    \begin{itemize}
    \item As shown in Table \ref{table:ablation_study_combined_single_column} and \ref{table:ablation_study_time}, RALAD demonstrated exceptional capability in bridging the real and simulated gap. Ensuring that the learned knowledge is not forgotten, the performance of all three models in simulation improved by approximately 10\%. Compared to using  the $l_2$ distance, RALAD performed better, with a relative improvement about 5\% in simulation. In addition, RALAD also achieved a significant reduction in training time. 
    \item In section \ref{sec:visual}, we clearly observed that the model’s performance significantly improved in simulation, after using RALAD. To further validate its effectiveness, we also used real autonomous driving car and conducted tests in parking lots. The results showed that the model’s performance in the real world also showed a remarkable boost, fully demonstrating the outstanding effectiveness of RALAD.
    \end{itemize}

    \section{CONCLUSIONS}
    \label{sec:conclusions}
    RALAD demonstrates a significant success in bridging the gap between real and simulated driving scenarios. By introducing RALAD, the model effectively constructs a mapping between the real and simulated environments. Experimental results show that RALAD not only maintains high accuracy in reality but also substantially improves detection performance in simulation. Additionally, the fine-tuning strategy employed by the model significantly enhances training efficiency. Overall, RALAD offers a more efficient and reliable solution for autonomous driving testing, advancing the development of autonomous driving technology. Future research could explore applying RALAD to a broader range of tasks and scenarios in autonomous driving.
    

    \section{ACKNOWLEDGEMENT}
    
    This work was supported in part by the National Natural Science Foundation of China (62072321), the Science and Technology Program of Jiangsu Province (BZ2024062), the Natural Science Foundation of the Jiangsu Higher Education Institutions of China (22KJA520007), Suzhou Planning Project of Science and Technology (2023ss03).


\begin{thebibliography}{99}

\bibitem{c1} Dong J, Chen S, Miralinaghi M, et al. Development and testing of an image transformer for explainable autonomous driving systems[J]. Journal of Intelligent and Connected Vehicles, 2022, 5(3): 235-249.

\bibitem{c2} Atakishiyev S, Salameh M, Yao H, et al. Explainable artificial intelligence for autonomous driving: A comprehensive overview and field guide for future research directions[J]. IEEE Access, 2024.

\bibitem{c3} Chib P S, Singh P. Recent advancements in end-to-end autonomous driving using deep learning: A survey[J]. IEEE Transactions on Intelligent Vehicles, 2023.

\bibitem{c4} Cui C, Ma Y, Cao X, et al. A survey on multimodal large language models for autonomous driving[C]//Proceedings of the IEEE/CVF Winter Conference on Applications of Computer Vision. 2024: 958-979.

\bibitem{c5} Mani K, Daga S, Garg S, et al. Monolayout: Amodal scene layout from a single image[C]//Proceedings of the IEEE/CVF Winter Conference on Applications of Computer Vision. 2020: 1689-1697.

\bibitem{c6} Yang W, Li Q, Liu W, et al. Projecting your view attentively: Monocular road scene layout estimation via cross-view transformation[C]//Proceedings of the IEEE/CVF conference on computer vision and pattern recognition. 2021: 15536-15545.

\bibitem{c7} Kim C, Kim U H. A Dual-Cycled Cross-View Transformer Network for Unified Road Layout Estimation and 3D Object Detection in the Bird's-Eye-View[C]//2023 20th International Conference on Ubiquitous Robots (UR). IEEE, 2023: 41-47.

\bibitem{c8} Alaba S Y, Ball J E. Deep learning-based image 3-d object detection for autonomous driving[J]. IEEE Sensors Journal, 2023, 23(4): 3378-3394.

\bibitem{c9} Wu H, Wen C, Li W, et al. Transformation-equivariant 3d object detection for autonomous driving[C]//Proceedings of the AAAI Conference on Artificial Intelligence. 2023, 37(3): 2795-2802.

\bibitem{c10} Hu Y, Yang J, Chen L, et al. Planning-oriented autonomous driving[C]//Proceedings of the IEEE/CVF Conference on Computer Vision and Pattern Recognition. 2023: 17853-17862.

\bibitem{c11} Geiger A, Lenz P, Stiller C, et al. Vision meets robotics: The kitti dataset[J]. The International Journal of Robotics Research, 2013, 32(11): 1231-1237.

\bibitem{c12} Sun P, Kretzschmar H, Dotiwalla X, et al. Scalability in perception for autonomous driving: Waymo open dataset[C]//Proceedings of the IEEE/CVF conference on computer vision and pattern recognition. 2020: 2446-2454.

\bibitem{c13} Caesar H, Bankiti V, Lang A H, et al. nuscenes: A multimodal dataset for autonomous driving[C]//Proceedings of the IEEE/CVF conference on computer vision and pattern recognition. 2020: 11621-11631.

\bibitem{c14} Bolte J A, Bar A, Lipinski D, et al. Towards corner case detection for autonomous driving[C]//2019 IEEE Intelligent vehicles symposium (IV). IEEE, 2019: 438-445.

\bibitem{c15} Breitenstein J, Termöhlen J A, Lipinski D, et al. Corner cases for visual perception in automated driving: some guidance on detection approaches[J]. arXiv preprint arXiv:2102.05897, 2021.

\bibitem{c16} Li K, Chen K, Wang H, et al. Coda: A real-world road corner case dataset for object detection in autonomous driving[C]//European Conference on Computer Vision. Cham: Springer Nature Switzerland, 2022: 406-423.

\bibitem{c17} Hu X, Li S, Huang T, et al. How simulation helps autonomous driving: A survey of sim2real, digital twins, and parallel intelligence[J]. IEEE Transactions on Intelligent Vehicles, 2023.

\bibitem{c18} Cui Y, Liu Z, Chen Y, et al. Retrieval-augmented multiple instance learning[J]. Advances in Neural Information Processing Systems, 2024, 36.

\bibitem{c19} Mondal A, Tigas P, Gal Y. Real2sim: Automatic generation of open street map towns for autonomous driving benchmarks[C]//Machine Learning for Autonomous Driving Workshop at the 34th Conference on Neural Information Processing Systems (NeurIPS). 2020.

\bibitem{c20}  Xiaozhi Chen, Kaustav Kundu, Ziyu Zhang, Huimin Ma, and Raquel Urtasun. Monocular 3d object detection for autonomous driving. In CVPR, 2016.

\bibitem{c21} Daoud A, Bunel C, Guériau M. CornerSim: A Virtualization Framework to Generate Realistic Corner-Case Scenarios for Autonomous Driving Perception Testing[J]. Procedia Computer Science, 2024, 238: 184-191.

\bibitem{c22} Shivam Kalra, Hamid R Tizhoosh, Charles Choi, Sultaan Shah, Phedias Diamandis, Clinton JV Campbell, and Liron Pantanowitz. Yottixel–an image search engine for large archives of histopathology whole slide images. Medical Image Analysis, 65:101757, 2020.

\bibitem{c23} Chengkuan Chen, Ming Y Lu, Drew FK Williamson, Tiffany Y Chen, Andrew J Schaumberg, and Faisal Mahmood. Fast and scalable search of whole-slide images via self-supervised deep learning. Nature Biomedical Engineering, 6(12):1420–1434, 2022.

\bibitem{c24} Anna Yeaton, Rahul G Krishnan, Rebecca Mieloszyk, David Alvarez-Melis, and Grace Huynh. Hierarchical optimal transport for comparing histopathology datasets. arXiv preprint arXiv:2204.08324, 2022.


\bibitem{c25} Udupa, Sumanth and Gurunath, Prajwal and Sikdar, Aniruddh and Sundaram, Suresh. MRFP: Learning Generalizable Semantic Segmentation from Sim-2-Real with Multi-Resolution Feature Perturbation. In CVPR, 2024.

\bibitem{c26}A. Kadian et al., “Sim2real predictivity: Does evaluation in simulation predict real-world performance?,” IEEERobot. Autom. Lett., vol. 5, no. 4, pp. 6670–6677, Oct. 2020.

\bibitem{c27} X. Hu, S. Li, T. Huang, B. Tang, R. Huai and L. Chen, "How Simulation Helps Autonomous Driving: A Survey of Sim2real, Digital Twins, and Parallel Intelligence," in IEEE Transactions on Intelligent Vehicles, vol. 9, no. 1, pp. 593-612, Jan. 2024

\bibitem{c28}Xie, Yiming et al. “SynFog: A Photo-realistic Synthetic Fog Dataset based on End-to-end Imaging Simulation for Advancing Real-World Defogging in Autonomous Driving.” ArXiv abs/2403.17094 (2024): n. pag.

\bibitem{c29} Marco Cuturi. Sinkhorn distances: Lightspeed computation of optimal transport. Advances in neural information processing systems, 26, 2013.

\bibitem{c30} Zengqun Zhao and Ziquan Liu and Yu Cao and Shaogang Gong and Ioannis Patras. AIM-Fair: Advancing Algorithmic Fairness via Selectively Fine-Tuning Biased Models with Contextual Synthetic Data.Proceedings of the Computer Vision and Pattern Recognition Conference (CVPR), June, 2025.

\bibitem{av4security} Baldini G. Testing and certification of automated vehicles (AV) including cybersecurity and artificial intelligence aspects[J]. EUR 30472 EN, JRC121631, 2020.

\bibitem{modelbysim} Yang Y, Chen D, Qin T, et al. E2e parking: Autonomous parking by the end-to-end neural network on the carla simulator[C]//2024 IEEE Intelligent Vehicles Symposium (IV). IEEE, 2024: 2375-2382.

\end{thebibliography}
\end{document}